\documentclass[10pt,twocolumn,twoside]{IEEEtran}

\usepackage{graphicx}
\usepackage{url}

\newcommand{\logp}{\ell}
\renewcommand{\sup}{\mbox{\sc sup}}
\renewcommand{\top}{\mbox{\sc top}}
\newcommand{\pop}{\mbox{\sc pop}}
\newcommand{\undef}{\mbox{\sc undef}}
\newcommand{\fastsubs}{\mbox{\sc fastsubs }}

\begin{document}

\title{FASTSUBS: An Efficient and Exact Procedure for Finding the Most Likely Lexical Substitutes Based on an N-gram Language Model}

\author{Deniz~Yuret
\thanks{
Copyright (c) 2012 IEEE. Personal use of this material is
permitted. However, permission to use this material for any other
purposes must be obtained from the IEEE by sending a request to
pubs-permissions@ieee.org.

D. Yuret is with the Department of Computer Engineering,
Ko\c{c} University, {\.I}stanbul, Turkey,
e-mail: {\tt dyuret@ku.edu.tr}.}}

\markboth{IEEE Signal Processing Letters,~Vol.~0, No.~0, January~0000}%
{Yuret: \fastsubs}

\maketitle

\begin{abstract}
Lexical substitutes have found use in areas such as paraphrasing, text
simplification, machine translation, word sense disambiguation, and
part of speech induction.  However the computational complexity of
accurately identifying the most likely substitutes for a word has made
large scale experiments difficult.  In this paper I introduce a new
search algorithm, \fastsubs, that is {\em guaranteed} to find the $K$
most likely lexical substitutes for a given word in a sentence based
on an n-gram language model.  The computation is sub-linear in both
$K$ and the vocabulary size $V$.  An implementation of the algorithm and a
dataset with the top 100 substitutes of each token in the WSJ section
of the Penn Treebank are available at \url{http://goo.gl/jzKH0}.
\end{abstract}

\begin{center} {\bfseries EDICS Category:} SPE-LANG \end{center}


\section{Introduction}

Lexical substitutes have proven useful in applications such as
paraphrasing \cite{mccarthy2007semeval}, text simplification
\cite{specia2012semeval}, and machine translation
\cite{mihalcea2010semeval}.  Best published results in unsupervised
word sense disambiguation \cite{Yuret2010coli}, and part of speech
induction \cite{yatbaz:2012:EMNLP} represent word context as a vector
of substitute probabilities.  Using a statistical language model to
find the most likely substitutes of a word in a given context is a
successful approach (\cite{Hawker2007,yuret2007ku}).  However the
computational cost of an exhaustive algorithm, which computes the
probability of every word before deciding the top $K$, makes large
scale experiments difficult.  On the other hand, heuristic methods run
the risk of missing important substitutes.

This paper presents the \fastsubs algorithm which can efficiently and
correctly identify the most likely lexical substitutes for a given
context based on an n-gram language model without going through most
of the vocabulary.  Even though the worst-case performance of
\fastsubs is still proportional to vocabulary size, experiments
demonstrate that the average cost is sub-linear in both the number of
substitutes $K$ and the vocabulary size $V$.  To my knowledge, this is
the first sub-linear algorithm that exactly identifies the top $K$
most likely lexical substitutes.

The efficiency of \fastsubs makes large scale experiments based on
lexical substitutes feasible.  For example, it is possible to compute
the top 100 substitutes for each one of the 1,173,766 tokens in the
WSJ section of the Penn Treebank \cite{marcus1999treebank} in under 5
hours on a typical workstation.  The same task would take about 6 days
with the exhaustive algorithm.  The Penn Treebank substitute data and
an implementation of the algorithm are available from the author's
website at \url{http://goo.gl/jzKH0}.

Section~\ref{sec:prob} derives substitute probabilities as defined by
an n-gram language model with an arbitrary order and smoothing.
Section~\ref{sec:algorithm} describes the \fastsubs algorithm.
Section~\ref{sec:correctness} proves the correctness of the algorithm
and Section~\ref{sec:complexity} presents experimental results on its
time complexity.  Section~\ref{sec:contributions} summarizes the
contributions of this paper.

\section{Substitute Probabilities}
\label{sec:prob}

This section presents the derivation of lexical substitute
probabilities based on an n-gram language model.  Details of this
derivation are important in finding an admissible algorithm that
identifies the most likely substitutes efficiently, without trying out
most of the vocabulary.

N-gram language models assign probabilities to arbitrary sequences of
words (or other tokens like punctuation etc.) based on their occurrence
statistics in large training corpora.  They approximate the
probability of a sequence of words by assuming each word is
conditionally independent of the rest given the previous $(n-1)$
words.  For example a trigram model would approximate the probability
of a sequence $abcde$ as:
\begin{equation} \label{eq:ngram}
p(abcde) = p(a) p(b|a) p(c|ab) p(d|bc) p(e|cd)
\end{equation}
\noindent where lowercase letters like $a$, $b$, $c$ represent words
and strings of letters like $abcde$ represent word sequences.  The
computation is typically performed using log probabilities, which
turns the product into a summation:
\begin{equation} \label{eq:logngram}
\logp(abcde) = \logp(a) + \logp(b|a) + \logp(c|ab) + \logp(d|bc) + \logp(e|cd)
\end{equation}
\noindent where $\logp(x) \equiv \log p(x)$.  The individual
conditional probability terms are typically expressed in back-off
form:\footnote{Even interpolated models can be represented in the
  back-off form and in fact that is the way SRILM stores them in ARPA
  (Doug Paul) format model files.}
\begin{equation}
\logp(c|ab) = \left\{ \begin{array}{ll}
\alpha(abc) & \mbox{if $f(abc)>0$}\\
\beta(ab) + \logp(c|b) & \mbox{otherwise}
\end{array} \right.\label{eq:backoff}
\end{equation}
\noindent where $\alpha(abc)$ is the discounted log probability
estimate for $\logp(c|ab)$ (typically slightly less than the log
frequency in the training corpus), $f(abc)$ is the number of times
$abc$ has been observed in the training corpus, $\beta(ab)$ is the
back-off weight to keep the probabilities add up to 1.  The formula
can be generalized to arbitrary n-gram orders if we let $b$ stand for
zero or more words.  The recursion bottoms out at unigrams (single
words) where $\logp(c)=\alpha(c)$.  If there are any out-of-vocabulary
words we assume they are mapped to a special $\langle\mbox{\sc
  unk}\rangle$ token, so $\alpha(c)$ is never undefined.

It is best to use both left and right context when estimating the
probabilities for potential lexical substitutes.  For example, in
\emph{``He lived in San Francisco suburbs.''}, the token \emph{San}
would be difficult to guess from the left context but it is almost
certain looking at the right context.  The log probability of a
substitute word given both left and right contexts can be estimated
as:
\begin{eqnarray} \label{eq:psub}
\logp(x|ab\_de) & \propto & \logp(abxde) \\
& \propto & \logp(x|ab) + \logp(d|bx) + \logp(e|xd) \nonumber
\end{eqnarray}

Here the ``$\_$'' symbol represents the position the candidate
substitute $x$ is going to occupy.  The first line follows from the
definition of conditional probability and the second line comes from
Equation~\ref{eq:ngram} except the terms that do not include the
candidate $x$ have been dropped.

The expression for the unnormalized log probability of a lexical
substitute according to Equation~\ref{eq:psub} and the decomposition
of its terms according to Equation~\ref{eq:backoff} can be combined to
give us Equation~\ref{eq:main}.  For arbitrary order n-gram models we
would end up with a sum of $n$ terms and each term would come from one
of $n$ alternatives.  

\begin{eqnarray} \label{eq:main}
\logp(x|ab\_de) &\propto& \mbox{\hspace*{5cm}}
\end{eqnarray}
\begin{eqnarray*}
& \left\{ \begin{array}{ll}
\alpha(abx) & \mbox{if $f(abx)>0$}\\
\beta(ab) + \alpha(bx) & \mbox{if $f(bx)>0$}\\
\beta(ab) + \beta(b) + \alpha(x) & \mbox{otherwise} \\
\end{array} \right. \\
+& \left\{ \begin{array}{ll}
\alpha(bxd) & \mbox{if $f(bxd)>0$}\\
\beta(bx) + \alpha(xd) & \mbox{if $f(xd)>0$}\\
\beta(bx) + \beta(x) + \alpha(d) & \mbox{otherwise} \\
\end{array} \right.\\
+& \left\{ \begin{array}{ll}
\alpha(xde) & \mbox{if $f(xde)>0$}\\
\beta(xd) + \alpha(de) & \mbox{if $f(de)>0$}\\
\beta(xd) + \beta(d) + \alpha(e) & \mbox{otherwise} \\
\end{array} \right.\\
\end{eqnarray*}

\section{Algorithm}
\label{sec:algorithm}

The task of \fastsubs is to pick the top $K$ substitutes ($x$) from a
vocabulary of size $V$ that maximize Equation~\ref{eq:main} for a
given context $ab\_de$.  Equation~\ref{eq:main} forms a tree where
leaf nodes are primitive terms such as $\beta(bx)$, $\alpha(xd)$, and
parent nodes are compound terms, i.e.  sums or conditional
expressions.  The basic strategy is to construct a priority queue of
candidate substitutes for Equation~\ref{eq:main} by composing
substitute queues for each of its sub-expressions.  The structure of
these queues and how they can be composed is described next, followed
by the construction of the individual queues for each of the
subexpressions.

\subsection{Upper bound queues}
\label{sec:queue}

A sum such as $\beta(bx) + \alpha(xd)$ is not necessarily maximized by
the $x$'s that maximize either of its terms.  What we can say for sure
is that the sum for any $x$ cannot exceed the upper bound $\beta(bx_1)
+ \alpha(x_2d)$ where $x_1$ maximizes $\beta(bx)$ and $x_2$ maximizes
$\alpha(xd)$.  We can find the $x$ that maximizes the sum by
repeatedly evaluating candidates until we find one whose value is (i)
larger than all the candidates that have been evaluated, and (ii)
larger than the upper bound for the remaining candidates.

Based on this intuition, we define an abstract data type called an
{\em upper bound queue} that maintains an {\em upper bound} on the
actual values of its elements.  Each successive {\em pop} from an
upper bound queue is not guaranteed to retrieve the element with the
largest value, but the remaining elements are guaranteed to have
values smaller than or equal to a non-increasing upper bound.  An
upper bound queue supports three operations:

\begin{itemize}
\item $\sup(q)$: returns an upper bound on the value of the
  elements in the queue.
\item $\top(q)$: returns the top element in the queue. Note that this
  element is not guaranteed to have the highest value.
\item $\pop(q)$: extracts and returns the top element in the queue and
  updates the upper bound if possible.
\end{itemize}

Upper bound queues can be composed easily.  Going back to our sum
example let us assume that we have valid upper bound queues $q_\alpha$
for $\alpha(xd)$ and $q_\beta$ for $\beta(bx)$.  The queue $q_\sigma$
for the sum $(\beta(bx)+\alpha(xd))$ has
$\sup(q_\sigma)=\sup(q_\alpha)+\sup(q_\beta)$ because the upper bound
for a sum clearly cannot exceed the total of the upper bounds for its
constituent terms.  $\top(q_\sigma)$ can return any element from the
queue without violating the contract.  However in order to find the
true maximum, we eventually need an element whose value exceeds the
upper bound for the remaining elements.  Thus we can bias our choice
for $\top(q_\sigma)$ to prefer elements that (i) have high values, and
(ii) reduce the upper bound quickly.  In practice
non-deterministically picking $\top(q_\sigma)$ to be one of
$\top(q_\alpha)$ or $\top(q_\beta)$ works well.  $\pop(q_\sigma)$ can
extract and return the same element from the corresponding child
queue.  If the upper bound of a child queue drops as a result, so does
the upper bound of the compound queue $q_\sigma$.

\subsection{Top level queue}
\label{sec:main}

The top level sum in Equation~\ref{eq:main} is a sum of $N$
conditional expressions for an order $N$ language model.  We can
construct an upper bound queue for the sum using the upper bound
queues for its constituent terms as described in the previous section.
Let $q$ represent the queue for the top level sum, $\delta \in C$
represent the constituent conditional expressions and $q_\delta$
represent their associated queues.  
\begin{eqnarray}
\sup(q) &=& \sum_{\delta \in C} \sup(q_\delta) \\
\top(q) &=& \top(q_\delta) \mbox{ for a random $\delta$.} \nonumber
\end{eqnarray}
For $\top(q)$ we non-deterministically pick the top element from one
of the children and $\pop(q)$ extracts and returns that same element
adjusting the upper bound if necessary.

As mentioned before $\top(q)$ does not necessarily return the element
with the maximum value.  In order to find the top $K$ elements
\fastsubs keeps popping elements from $q$ and computes their true
values according to Equation~\ref{eq:main} until at least $K$ of them
have values above the upper bound for the remaining elements in the
queue.  Table~\ref{tab:pseudocode} gives the pseudo-code for
\fastsubs.

\begin{table}[ht]\centering\fbox{
\begin{minipage}{.8\linewidth}
\noindent\fastsubs($S$, $K$)\begin{enumerate}
\item Initialize upper bound queue $q$ for context $S$.
\item Initialize set of candidate words $X=\{\}$.
\item {\sc while} $|\{x: x \in X, \logp(x|S) \geq \sup(q)\}| < K$\\
{\sc do} $X := X \cup \{\pop(q)\}$
\item Return top $K$ words in $X$ based on $\logp(x|S)$.
\end{enumerate}
\end{minipage}
}
\caption{Pseudo-code for \fastsubs.  Given a word context $S$ and the
  desired number of substitutes $K$, \fastsubs returns the set of top
  $K$ words that maximize $\logp(x|S)$.}
\label{tab:pseudocode}
\end{table}

This procedure will return the correct result as long as $\pop(q)$
cycles through all the words in the vocabulary and the upper bound for
the remaining elements, $\sup(q)$, is accurate.  The loop can in fact
cycle through all the words in the vocabulary because at least one of
the subexpressions, $\alpha(x)$, is well defined for every word.  The
accuracy of $\sup(q)$ depends on the accuracy of the upper bounds for
constituent terms, which are described next.

\subsection{Queues for conditional expressions}

Conditional expressions indicated by ``\{'' in Equation~\ref{eq:main}
pick their topmost child whose $\alpha$ argument has been observed in
the training corpus.  Let $q_\delta$ be the queue for such a
conditional expression and $\sigma \in C_\delta$ be its children
terms.  Let ${\sigma}_{\max} = \arg\max_{\sigma \in C_\delta}
\sup(q_\sigma)$ be the child whose queue has the maximum upper bound.
The upper bound for $q_\delta$ cannot exceed the upper bound for
$q_{{\sigma}_{\max}}$ because the value of the conditional expression
for any given $x$ is equal to the value of one of its children.  Thus
we define the queue operations for conditional expressions based on
$q_{{\sigma}_{\max}}$:

\begin{eqnarray}
  \sup(q_\delta) &=& \sup(q_{{\sigma}_{\max}}) \\
  \top(q_\delta) &=& \top(q_{{\sigma}_{\max}}) \nonumber
\end{eqnarray}

\subsection{Queues for sums of primitive terms}

As described in Section~\ref{sec:queue}, the upper bound of a queue
for a sum like $\beta(bx)+\alpha(xd)$ is equal to the sum of the upper
bounds of the constituent queues.  It turns out that for sums of
primitive terms, only the $\alpha$ term that has the candidate word
$x$ as an argument has a non-constant upper-bound.  The language model
defines $\beta$ to be 0 for any word sequence that does not appear in
the training set.  Therefore the $\beta$ terms that have the candidate
word $x$ as an argument always have the upper bound 0.  Finally, the
$\alpha$ and $\beta$ terms without the candidate word $x$ act as
constants.

For notational consistency we define upper bounds for the constant
terms as well.  Let $A$ and $B$ represent sequences of zero or more
words that do not include the candidate $x$.  We have:
\begin{eqnarray}
  \sup(q_\alpha(A)) &=& \alpha(A) \\
  \sup(q_\beta(B)) &=& \beta(B) \nonumber
\end{eqnarray}

For $\beta$ terms with $x$ in their argument, many words from the
vocabulary would be unobserved in the argument sequence and share the
maximum $\beta$ value of 0.  In the rare case that all vocabulary
words have been observed in the argument sequence, they would each
have negative $\beta$ values and 0 would still be a valid upper bound.
Thus \fastsubs uses the constant 0 as an upper bound for $\beta$ terms
with $x$.
\begin{eqnarray}
  \sup(q_\beta(AxB)) &=& 0
\end{eqnarray}

Only the $\alpha$ term with an $x$ argument has an upper bound queue
as described in the next section.  \fastsubs picks the top element for
a sum of primitive terms only from its $\alpha$
constituent.\footnote{
  Remember that the top value in an upper bound queue is not
  guaranteed to have the largest value.  Thus ignoring the $\beta$
  terms does not effect the correctness of the algorithm.}
Let $q_\sigma$ be the queue for a sum of primitive terms and let $\gamma
\in C_\sigma$ indicate its constituents ($\alpha$, $\beta$, constant
or otherwise).  We have:
\begin{eqnarray}
\sup(q_\sigma) &=& \sum_{\gamma \in C_\sigma} \sup(q_\gamma)\\
\top(q_\sigma) &=& 
\left\{ \begin{array}{l}
\top(q_\alpha) \mbox{ if the $\alpha$ term has an $x$ argument.}\\
\undef \mbox{ otherwise.}
\end{array} \right. \nonumber
\end{eqnarray}

\subsection{Queues for primitive terms}
\label{sec:primitive}

\fastsubs pre-computes actual priority queues (which satisfy the upper
bound queue contract) for $\alpha$ terms that include $x$ in their
argument:
  \begin{eqnarray}
    \sup(q_\alpha(AxB)) &=& \max_x \alpha(AxB) \\
    \top(q_\alpha(AxB)) &=& \arg\max_x \alpha(AxB) \nonumber
  \end{eqnarray}
Here $A$ and $B$ stand for zero or more words and $x$ is a candidate
lexical substitute word.  $\sup(q_\alpha)$ gives the real maximum,
thus provides a tight upper bound.  $\top(q_\alpha)$ is guaranteed
to return the element with the highest value.

The $q_\alpha$ queues are constructed once in the beginning of the
program as sorted arrays and re-used in queries for different
contexts.  The construction can be performed in one pass through the
language model and the memory requirement is of the same order as the
size of the language model.  Candidates that have not been observed in
the argument context will be at the bottom of this queue because
$\alpha(AxB)\equiv-\infty$ if $f(AxB)=0$.  To save memory such $x$ are
not placed in the queue.  Thus after we run out of elements in
$q_\alpha$ the queue returns:
\begin{eqnarray}
   \sup(q_\alpha(AxB)) &=& -\infty \\
   \top(q_\alpha(AxB)) &=& \undef \nonumber
\end{eqnarray}

\section{Correctness} 
\label{sec:correctness}

As mentioned in Section~\ref{sec:algorithm}, the correctness of the
algorithm depends on two factors: (i) the $\sup(q)$ function should
return an upper bound on the remaining values in $q$, and (ii) the
$\pop(q)$ function should cycle through the whole vocabulary for the
top level queue.

The correctness of the $\sup(q)$ function can be proved recursively.
For primitive terms $\sup(q)$ is equal to the actual maximum (e.g. for
$q_\alpha$), or is an obvious upper bound
(e.g. $\sup(q_\beta(AxB))=0$).  For sums, $\sup(q)$ is equal to the
sum of the upper bounds for the children and, for conditional
expressions, $\sup(q)$ is equal to the maximum of the upper bounds for
the children.

To prove that $\pop(q)$ will cycle through the entire vocabulary it
suffices to show that the queue for at least one child of $q$ will
cycle through the entire vocabulary.  This is in fact the case because
one of the children will always include the term $\alpha(x)$ whose
queue contains the entire vocabulary.

\section{Complexity}
\label{sec:complexity}

A exhaustive algorithm to find the most likely substitutes in a given
context could try each word in the vocabulary as a potential
substitute $x$ and compute the value of the expression given in
Equation~\ref{eq:main}.  The computation of Equation~\ref{eq:main}
requires $O(N^2)$ operations for an order $N$ language model, which we
will assume to be a constant.  If we have $V$ words in our vocabulary
the cost of the exhaustive algorithm to find a single most likely
substitute would be $O(V)$.

\begin{figure}[ht]\centering
\includegraphics[width=\linewidth]{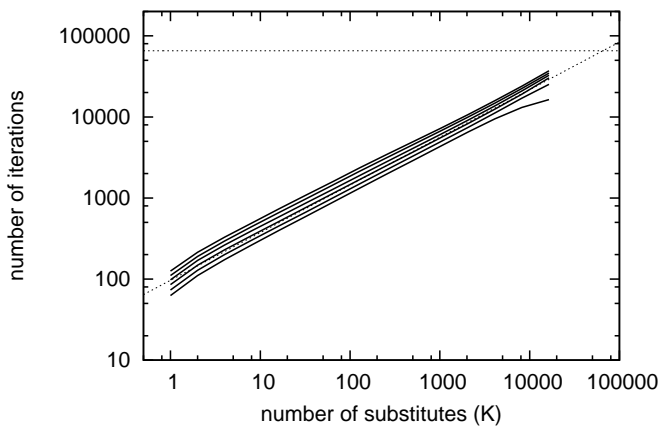}
\caption{Number of iterations as a function of the number of
  substitutes $K$ and the vocabulary size $V$.  The solid curves
  represent results with vocabulary sizes from 16K to 512K.  The
  horizontal dotted line gives the cost of the exhaustive algorithm
  for $V=64K$.  The diagonal dotted line is a functional approximation
  in the form $K^\lambda V^{(1-\lambda)}$ for $V=64K$ and
  $\lambda=0.5878$.}
\label{fig:fastsubs}
\end{figure}

In order to quantify the efficiency of \fastsubs on a real world
dataset, I used a corpus of 126 million words of WSJ data as the
training set and the WSJ section of the Penn Treebank
\cite{marcus1999treebank} as the test set.  Several 4-gram language
models were built from the training set using Kneser-Ney smoothing in
SRILM \cite{stolcke2002srilm} with vocabulary sizes ranging from 16K
to 512K words.  The average number of $\pop(q)$ operations for the top
level upper bound queue was measured for number of substitutes $K$
ranging from 1 to 16K.  Figure~\ref{fig:fastsubs} shows the results.

The time cost of \fastsubs depends on the number of iterations of the
while loop in Table~\ref{tab:pseudocode} which in turn depends on the
quality of words returned by $\pop(q)$ and the tightness of the upper
bound given by $\sup(q)$.  The worst case is no better than the
exhaustive algorithm's $O(V)$.  However Figure~\ref{fig:fastsubs}
shows that the average performance of \fastsubs on real data is
significantly better when $K \ll V$.  The number of $\pop(q)$
operations in the while loop to get the top $K$ substitutes is
sub-linear in $K$ (the slope of the log-log curves are around 0.5878)
and approaches the vocabulary size $V$ as $K \rightarrow V$.  The
effect of vocabulary size is practically insignificant: increasing
vocabulary size from 16K to 512K less than doubles the average number
of steps for a given $K$.

As a practical example, it is possible to compute the top 100
substitutes for each one of the 1,173,766 tokens in Penn Treebank with
a vocabulary size of 64K in under 5 hours on a typical 2012
workstation.\footnote{
Running a single thread on an Intel Xeon E7-4850 2GHz processor.}
The same task would take about 6 days for the exhaustive algorithm.

\section{Contributions}
\label{sec:contributions}

Finding likely lexical substitutes has a range of applications in
natural language processing.  In this paper we introduced an exact and
efficient algorithm, \fastsubs, that is guaranteed to find the $K$
most likely substitutes for a given word context from a $V$ word
vocabulary.  Its average runtime is sub-linear in both $V$ and $K$
giving a significant improvement over an exhaustive $O(V)$ algorithm
when $K \ll V$.  An implementation of the algorithm and a dataset with
the top 100 substitutes of each token in the WSJ section of the Penn
Treebank are available at \url{http://goo.gl/jzKH0}.

\section*{Acknowledgments}

I would like to thank the members of the Natural Language Group at
USC/ISI for their hospitality and for convincing me that a \fastsubs
algorithm is possible.

\bibliographystyle{IEEEtran}
\bibliography{fastsubs}
\end{document}